\documentclass[conference]{IEEEtran}
\IEEEoverridecommandlockouts
\usepackage{cite}
\usepackage{amsmath,amssymb,amsfonts}
\usepackage{algorithmic}
\usepackage{graphicx}
\usepackage{textcomp}
\usepackage{xcolor}
\usepackage{multirow}
\begin{document}

\title{Skeleton-based Human Action Recognition via Convolutional Neural Networks (CNN) }

\author{\IEEEauthorblockN{Ayman Ali, Ekkasit Pinyoanuntapong, Pu Wang, Mohsen Dorodchi}
\IEEEauthorblockA{\textit{College of Computing and Informatics} \\
\textit{University of North Carolina at Charlotte}\\
Charlotte, United States \\
aali26, epinyoan, pwang13, mdorodch@uncc.edu}
}
\graphicspath{{images/}}
\maketitle

\begin{abstract}
Recently, there has been a remarkable increase in the interest towards skeleton-based action recognition within the research community, owing to its various advantageous features, including computational efficiency, representative features, and illumination invariance. Despite this, researchers continue to explore and investigate the most optimal way to represent human actions through skeleton representation and the extracted features. As a result, the growth and availability of human action recognition datasets have risen substantially. In addition, deep learning-based algorithms have gained widespread popularity due to the remarkable advancements in various computer vision tasks. Most state-of-the-art contributions in skeleton-based action recognition incorporate a Graph Neural Network (GCN) architecture for representing the human body and extracting features. Our research demonstrates that Convolutional Neural Networks (CNNs) can attain comparable results to GCN, provided that the proper training techniques, augmentations, and optimizers are applied. Our approach has been rigorously validated, and we have achieved a score of $95\%$ on the NTU-60 dataset.
\end{abstract}

\section{\textbf{Introduction}}
Recently, the research community has been witnessing a surging interest in skeleton-based action recognition, owing to its many advantageous attributes such as computational efficiency, informative features, and immunity to changes in lighting conditions. The field continues to explore the optimal way of representing human actions through skeleton representation and feature extraction. This has resulted in a marked increase in the number of human action recognition datasets. On the other hand, deep learning-based algorithms have gained immense popularity as a result of their effectiveness in various computer vision tasks. Most cutting-edge contributions in the realm of skeleton-based action recognition adopt Graph Neural Network (GCN) architecture for representing the human body's articulated structure and extracting features. However, our findings demonstrate that Convolutional Neural Networks (CNNs) can deliver comparable results to GCN if proper training methods, data augmentation techniques, and the appropriate optimizer are utilized.

The prospect of equipping machines with human-like visual capabilities has garnered significant attention among researchers, inspiring the development of various technologies, algorithms, and techniques to facilitate this task. To date, numerous computer vision tasks have been successfully addressed, including image classification \cite{krizhevsky2012imagenet} and object detection \cite{ren2015faster}. Among the active research areas in computer vision is human activity analysis in videos, including human action detection, recognition, and prediction.

The human gesture, which is characterized by a shorter duration and less complex movements performed by a limited number of body parts, represents the first category of human action. Examples of human gestures include hand waving and head nodding. On the other hand, human action encompasses longer-duration movements involving more body parts. A sequence comprising multiple actions is referred to as human activity. Finally, human interaction contains human interactions with the surrounding environment, including both human-to-human and human-to-object interactions. The presence of multiple humans or interactions with various objects increases the complexity of motion analysis, which can further be complicated by online or offline human behavior analysis.

In the past, capturing human actions through human-performance systems often necessitated the application of markers on the subjects' bodies and the use of distinctive attire. Despite overcoming these limitations, the high cost of cameras remained a hindrance to widespread adoption \cite{barker2006accuracy}. However, recent advancements have led to the development of cost-effective contactless sensing devices, such as Microsoft Kinect, Intel Realsense, and Doppler radar. Historically, the RGB modality in human action capture has been challenged by various factors, including illumination, occlusions, background clutter, frame rate, viewpoint, and biometric variations. By contrast, RGB-D sensors have mitigated some of these difficulties, particularly with regard to illumination, and provide a crucial advantage by enabling the generation of 3D structural information of the scene. As a result, estimating 3D human skeleton joints has become a relatively straightforward process in the RGB-D modality.

For many years, machine learning algorithms have been the primary technologies incorporated to address various computer vision problems, including action recognition. The conventional approach to human motion analysis involves capturing and representing spatiotemporal information, which enhances the accuracy and robustness of the analysis. Typically, features are extracted manually and incorporated into classical machine learning algorithms to perform different tasks. The research community has explored various features representations, including joint coordinates \cite{huang2020long}, the center of gravity \cite{raj2020exploring}, the angle between skeleton joints \cite{huynh2009metrics}, motion velocity \cite{yan2019convolutional}, and co-occurrence features \cite{li2018co}. The selection of the appropriate algorithm also recreates a vital role in the machine learning era. Many algorithms have been utilized, including Support Vector Machine (SVM) \cite{boser1992training}, Linear Discriminant Analysis (LDA) \cite{liu2017fusing}, Naive Bayes Nearest Neighbor \cite{weng2017spatio}, Logistic Regression \cite{tang2014human}, and KNN \cite{ubalde2016skeleton}. However, the generalization of machine learning algorithms is challenging and requires significant effort in feature engineering.

Recent advancements in deep learning-based techniques have demonstrated superior performance in various computer vision problems, such as image classification \cite{krizhevsky2012imagenet}, object detection \cite{ren2015faster}, action recognition \cite{duan2021revisiting}, and action detection \cite{wang2021proposal}. The increasing interest in skeleton-based representation has been driven by the abundant and discriminative features that can be derived from skeleton joints. For instance, features such as Skeleton Map \cite{du2015skeleton}, Joint Trajectory Map \cite{wang2016action}, Joint Distance Map \cite{li2017joint}, and many others can be extracted from the skeleton joint data alone. Consequently, the utilization of deep-learning algorithms has emerged as a prevalent approach for skeleton-based human action recognition.
Our contributions are summarized as follows:

\begin{itemize} \item Construct an easy-to-integrate and modular Convolutional Neural Network (CNN) for the action recognition task, which attains results comparable to the State-of-the-Art (SOTA) methods.
 \item Despite the prevalence of graph neural network-based methods in the SOTA contributions to the action recognition task, we demonstrate that CNNs can attain comparable results by implementing various training techniques.
\item Our results indicate that incorporating a diverse set of augmentation techniques leads to an improvement in the generalization and robustness of the model.
\item Our findings reveal that utilizing a margin-based cosine loss function instead of the conventional cross-entropy loss leads to a significant enhancement in performance. \end{itemize}

\section{\textbf{Related Work}}
Deep-learning approaches have been demonstrated superiority over conventional machine learning algorithms in various computer vision tasks, including image classification as demonstrated in \cite{krizhevsky2012imagenet} in the ImageNet dataset, object detection as introduced by Ren et al. \cite{ren2015faster} with their Faster R-CNN framework, action recognition as recently revisited by Duan et al \cite{duan2021revisiting}, and action detection as proposed by Wang et al. \cite{wang2021proposal}. In recent years, skeleton-based representations have garnered increasing attention due to the wealth of rich and discriminative features that can be obtained from the skeleton joints. Examples of such representations include the Skeleton Map\cite{du2015skeleton}, Joint Trajectory Map\cite{wang2016action}, Joint Distance Map \cite{li2017joint}, among others, all of which are based solely on features extracted from the skeleton joints data.

\subsubsection{\textbf{Convolution Neural Network - CNN Approaches}}
Wang \textit{et al.} \cite{wang2016action} proposed the Joint Trajectory Map (JTM), a representation that captures the spatiotemporal information of a sequence in the form of 2D images. This work emphasizes human motion magnitude and speed as the core features, represented by the saturation value in the HSV (Hue-Saturation-Value) color space, with higher motion magnitude and speed yielding a higher saturation value. Du \textit{et al.} \cite{du2015skeleton} introduced a method for encoding the spatiotemporal features of an action sequence into an image matrix. This representation is constructed by vertically encoding the skeleton's joint coordinates into the RGB channels and horizontally encoding the sequence frames. The encoded information is then quantified, normalized, and transformed into an image for use with a CNN classification network. In addition to the skeleton map representation proposed by Du \textit{et al.}, Li \textit{et al.} \cite{li2017skeleton} proposed a two-stream CNN-based network that leverages the skeleton map. Li \textit{et al.} \cite{li2017joint} extracted discriminative features from the pair-wise distances between skeleton joints in a human action sequence to construct a Joint Distance Map (JDM), which maps the skeleton sequence into images. Bilinear interpolation was applied to address the issue of variable sequence duration. Li \textit{et al.} \cite{li2018co} embarked on the problem of joint co-occurrence by proposing an end-to-end hierarchical framework that facilitates better feature learning, gradually aggregating point-level features into global co-occurrence features. Ke \textit{et al.} \cite{ke2017skeletonnet} proposed a translation, scale, and rotation invariant body part vector-based representation, using geometrical features of the skeleton to generate two sets of features: cosine distance and normalized magnitude. The human skeleton joints are grouped into five parts: trunk, right arm, left arm, right leg, and left leg. Both cosine distance and normalized magnitude features are computed within the body and skeleton, yielding ten feature vectors fed into a CNN for further feature extraction and training.

\subsubsection{\textbf{Recurrent Neural Network - RNN Approaches}}
The recognition of human actions is a problem of spatiotemporal representation. To capture the discriminative features, researchers have proposed utilizing conventional RNN networks. Yet, the vanishing gradient problem has necessitated adapting memory gating techniques, such as long-short term memory (LSTM) and gated recurrent unit (GRU). To address this issue, Du \textit{et al.} \cite{du2015hierarchical} proposed a hierarchical bidirectional recurrent neural network (BRNN) that models the long-term temporal sequence through an end-to-end approach. To facilitate better feature representation, the human body is divided into five parts and processed through five different BRNN subnets. The outputs are fused and classified in higher layers before being fed into the final BRNN. To further improve the system's robustness, Du \textit{et al.} \cite{du2016representation} implemented random rotation transformation during preprocessing and scale transformation to account for varying human sizes.
Salient motion patterns have been leveraged as essential features in various human action-related tasks. However, LSTM struggles to capture the spatiotemporal dynamics. To resolve this, Veeriah \textit{et al.} \cite{veeriah2015differential} proposed a differential gating for LSTM-RNN networks that quantifies the salient motion between frames through the derivative of state (DoS).
Skeleton-based action recognition presents several challenges, including the intra-frame joint spatiotemporal information of all frames in a sequence. To address these challenges, Sogn \textit{et al.} \cite{song2017end} proposed an end-to-end spatiotemporal attention model that adaptively learns intra-frame and inter-frame dependencies through recurrent RNN and LSTM.
Human skeleton data contains discriminative features such as acceleration, angles, spatial coordinates, orientation, and velocity. Zhang \textit{et al.} \cite{zhang2017geometric} evaluated eight geometric features on a 3-layer LSTM network and found that computing the distance between joints and selected lines outperformed the other features.

\subsubsection{\textbf{Graph Convolution Network - GCN Approaches}}
Yan \textit{et al.} \cite{yan2018spatial} introduced the initial spatiotemporal graph convolutional network (ST-GCN), which established edges between joints in both the intra-frame and inter-frame dimensions. Typically, in graph-based techniques, the convolution operation is performed over the neighboring joints within the receptive field. However, conducting a convolution over non-Euclidean data necessitates a judicious partition strategy to attain a label map.
Although ST-GCN \cite{yan2018spatial} achieved substantial results in the human action recognition task, it faced several limitations. The graph representation in ST-GCN was created based on the human body's physical structure, which may not be ideal for characterizing human action. For example, in an act like clapping, the relationship between the hands is crucial in classifying the action. In this case, ST-GCN fails to capture such a relationship since the two hands are not connected and are distant from each other in the kinematic tree. Additionally, certain geometric features, such as bone orientation and length, are challenging to represent in graph-based algorithms. To overcome these limitations, Shi \textit{et al.} \cite{shi2020skeleton} proposed an end-to-end data-driven multi-stream attention-enhanced adaptive graph convolutional network (MS-AAGCN), where the graph topology is learned and generated adaptively from the input data.
In practical scenarios, occlusion is a prevalent challenge that is often unavoidable. Song \textit{et al.} \cite{song2019richly} proposed a GCN-based model that is trained on incomplete spatiotemporal skeleton data, with intra-frame joints and inter-frame frames masked to imitate occlusion effects. Moreover, Yoon \textit{et al.} \cite{yoon2021predictively} incorporated noise into the skeleton data to enhance the model's robustness.

\section{\textbf{Data Preprocessing}}
Typically, skeleton data is represented in a camera coordinate system, leading to a diverse representation of captured sequences from different viewpoints, as illustrated in Figures \ref{fig:skeleton_representation} (A), (B), and (C). To mitigate the impact of view variation, a common approach is to transform the skeleton data into a unified coordinate system \cite{de2020infrared, lee2017ensemble, rahmani2017learning, zhang2019view}. This is typically accomplished through a sequence of geometric translation, rotation, and normalization operations, as depicted in Figures \ref{fig:skeleton_representation} (D), (E), and (F).
There are various transformation strategies presented in the literature. In frame-based approaches, the transformation operation is applied to each frame in the sequence, which, however, results in the loss of some relative motions. For example, when applied to the walking action, this technique produces an effect as if walking on a treadmill. On the other hand, in sequence-based strategies, the first frame in the sequence is designated as the reference frame, and all transformations applied to the subsequent frames are relative to this reference frame. This approach provides a more realistic representation of human skeleton motions.

\begin{figure}[ht]
\includegraphics[width=1.0\linewidth]{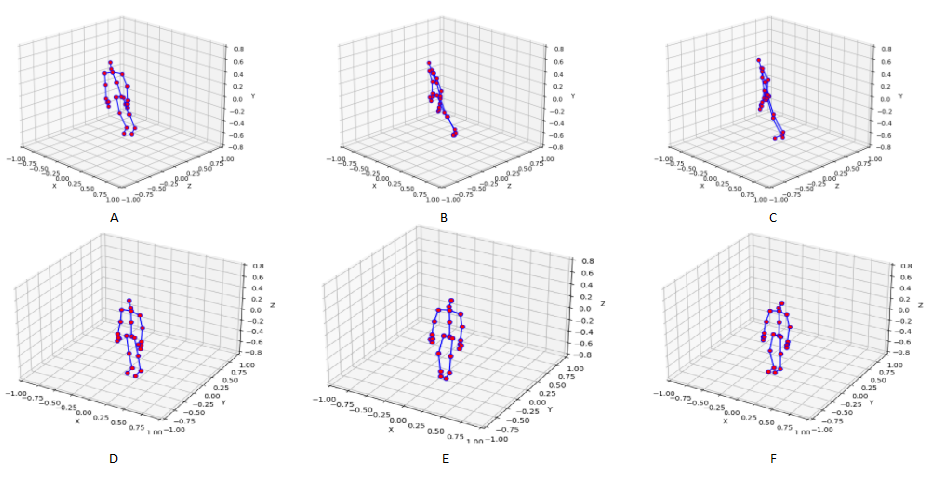}
\caption{Action representation from NTU-D 60 dataset A) -45\textdegree  skeleton visualization, B) 0 \textdegree  skeleton visualization, C) 45\textdegree  skeleton visualization. (D, E, F) are the transformed skeleton for the same skeletons in (A, B, C)}
\label{fig:skeleton_representation}
\end{figure}
\subsection{Encoding Skeleton to Image}
Given a sequence of skeletons $S$, where $s \in (1, ..., S)$, and the $j^{th}$ joint in the $t^{th}$ frame of skeleton $s$ is represented as $s_{t, j} = [x_{t, j}, y_{t, j}, z_{t, j}]$, with $j \in (1, ..., J)$ denoting the number of joints in a frame and $t \in (1, ..., T)$ denoting the total number of frames in the sequence $S$.

\begin{figure}[ht]
\includegraphics[width=1.0\linewidth]{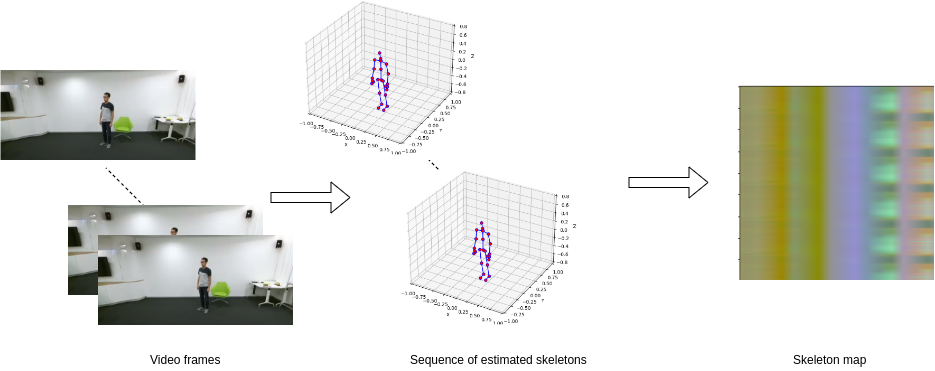}
\caption{The pipeline of generating the skeleton map image}
\label{fig:generating_skeleton}
\end{figure}

Inspired by the work of Du et al. \cite{du2015skeleton}, we transform the raw skeleton data into a skeleton map image that preserves spatiotemporal information, as depicted in Fig. \ref{fig:generating_skeleton}. Given an RGB image of dimensions $[H, W, C]$, where $H$ represents height, $W$ represents width, and $C$ represents the conventional RGB image channels, we map the action sequence $S$, represented in the dimension of $T \times N \times 3$, to an image of dimensions $[H, W, C]$. Here, $T$ represents the number of frames and maps to the height of the image, $N$ represents the number of joints and maps to the width of the image, and the last dimension represents the 3D joint coordinates of the joints in a frame, which are mapped to the three channels of the image.

Since raw skeleton data may have different value ranges than the image, pixel normalization is necessary to ensure that the mapped values are in the range of $0-255$. This is accomplished by:

\begin{equation}
P_{t,j} = floor(255 \times \frac{s_{t, j} - C_{min}}{C_{max}-C_{min}})
\end{equation}

\section{\textbf{Data Augmentation}}
Data augmentation is a well-established technique utilized to enhance the performance of machine learning algorithms by diversifying the training data. This approach involves synthesizing new samples from the existing training data by employing transformations such as scaling, rotation, translation, and other deformations. Adding these transformed samples to the training set can improve the generalization performance of the machine learning algorithm, making it more resilient to variations in the input data. Data augmentation is widely applied across several deep-learning tasks, including image classification, object detection, and natural language processing.

On the other hand, Skeleton-based data augmentation is used explicitly for tasks involving 3D pose information, such as 3D human pose estimation. This approach involves transforming the 3D joint positions of a skeleton to generate new pose samples. Image-based data augmentation is utilized for tasks where the input data consists of images, including image classification and object detection.

In this research, we investigated various image-based and skeleton-based techniques for data augmentation, as outlined in TABLE. Our approach is inspired by RandAugmentation \cite{cubuk2020randaugment}, which randomly injects the training pipeline with predefined augmentations based on the number and magnitude of augmentations to be applied.

The \textbf{Flipping Sequence} augmentation technique, as described in \cite{rao2021augmented}, involves the generation of synthetic pose sequences by horizontally flipping the input pose sequences. This is achieved through the reflection of the 3D joint positions across a horizontal plane when applied to skeleton data and by flipping in both the horizontal and vertical planes when applied to image data. As a result, a new pose sequence with actions performed in the opposite direction is produced.

The \textbf{Geometric Rotation} augmentation, as discussed in \cite{cubuk2020randaugment}, generates new synthetic samples by applying random rotations to the input data. This is achieved by rotating the input images or 3D objects around a fixed point, utilizing a specified angle and center of rotation.

The \textbf{Cutout} augmentation, as presented in \cite{devries2017improved}, is a regularization technique utilized in deep learning. This technique involves randomly masking a portion of an image during the training process, which necessitates the model to learn to ignore these regions and focus on the remaining parts. This can enhance the model's generalization to new data, thus improving its performance. Cutout augmentation is frequently adopted in image classification tasks, where it can aid the model in recognizing objects in images, despite variations in their appearance. It can be easily implemented in most deep learning frameworks as a simple and effective regularization method for deep learning models.

The \textbf{Zoom} augmentation, as described in\cite{cubuk2020randaugment}, involves rescaling the input images or 3D objects using a specified zoom factor, which determines the extent of the zooming. This augmentation method is appropriate for both image and skeleton data. Additionally, the \textbf{Shear} augmentation, as discussed in Cubuk et al. (2020)\cite{cubuk2020randaugment}, is achieved through the utilization of a linear transformation that maps the x-axis or y-axis coordinates of the input data to new positions. The \textbf{Translate} augmentation, also presented in Cubuk et al. (2020)\cite{cubuk2020randaugment}, involves the shifting of the x-axis, y-axis, or both coordinates of the input data by a specified amount.

The introduction of various types of noise to image data during training has been demonstrated to enhance the generalizability and robustness of machine learning models \cite{shorten2019survey}. \textbf{Salt-and-Pepper} noise \cite{liu2022towards} is a form of noise that mimics the effect of corrupted or missing pixels in the input data. This is achieved by randomly setting a specified proportion of pixels to either the minimum or maximum intensity value, creating the appearance of salt and pepper grains on the image. The remaining pixels remain unaltered.

Another noise-based augmentation strategy is \textbf{Localvars} noise \cite{shorten2019survey}. This method generates random noise samples from a specified distribution. \textbf{Speckle} noise \cite{shorten2019survey} generates random noise samples from a Poisson distribution with a specified mean. In contrast, \textbf{Gaussian} noise \cite{shorten2019survey} generates random noise samples from a Gaussian distribution with a specified mean and standard deviation.

In the context of skeleton data, \textbf{Bone Shuffling} augmentation is applied. This is achieved by randomly permuting the 3D joint positions of the skeleton, resulting in a new pose sequence in which a different body configuration performs the same actions. Additionally, \textbf{Bone Masking} augmentation is implemented by setting the 3D joint positions of the skeleton to zero for a randomly selected subset of bones. A similar technique can also be applied on the frame level rather than the bone level.

\begin{figure}[ht]
\includegraphics[width=1.0\linewidth]{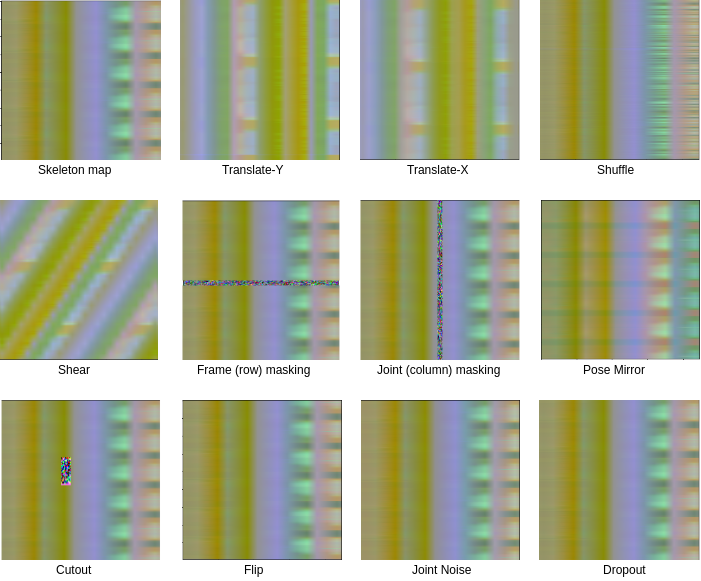}
\caption{Various augmentation implementation}
\label{fig:skeleton_representation}
\end{figure}

\section{Loss function}
In the realm of Deep Learning, a loss function serves as an evaluation metric of a model's ability to predict the desired outcome. The objective of training a model is to find the optimal set of parameters that minimize the loss function, leading to the model's improved accuracy in making predictions on unseen data. The use of classification loss is prevalent in classifying outputs into discrete class labels. The Cross-Entropy loss is a popular choice for action recognition classification problems \cite{suarez2021tutorial}. Although the Cross-Entropy loss often proves effective in training models, it has certain limitations, including sensitivity to the relative magnitude of predicted and true values, rendering it challenging to use in specific scenarios.

One critical observation when utilizing the Cross-Entropy loss function is that the distance between samples of different classes is minimal. As a result, the classifier trained with Cross-Entropy is susceptible to making incorrect predictions. An intuitive alternative to the Cross-Entropy loss is the implementation of metric learning techniques, which involve learning a distance metric from the data. Unlike conventional Machine Learning, where the aim is to learn a mapping function between inputs and outputs, the goal of metric learning is to learn a metric that can be utilized to measure the distance between data points \cite{suarez2021tutorial}. This distance metric can then be leveraged for making predictions or other tasks, maximizing the inter-class distance and minimizing the intra-class distance.
The Additive Angular Margin Loss (AAML) \cite{deng2019arcface} is a widely used loss function in Deep Learning for face recognition. It is designed to learn a discriminative feature representation for face images by maximizing the angular margin between the features of the same identity while minimizing the angular margin between the features of different identities. The AAML loss function is based on learning a weight vector and a feature vector for each identity. The angle between the weight and feature vectors is maximized for the same identity and minimized for different identities. The loss function encourages the weight and feature vectors to lie on the surface of a hypersphere. The large margin between the vectors of the same and different identities enhances the learned feature representation's discriminative power. The AAML loss function has been demonstrated to outperform other loss functions for face recognition on several benchmark datasets, resulting in a 1.5\% improvement in classifier accuracy.

\section{Deep-Learnig Optimizer and Schedulers}
\subsection{Training Optimizers}
The purpose of an optimizer in deep learning is to determine the set of parameters that minimize the loss function, thus yielding the model configuration most representative of the data. This is achieved through iterative refinement of the model parameters, guided by various algorithms and techniques, with the optimizer making progressive adjustments to the parameters until an optimal solution is reached. Commonly used optimizers include Stochastic Gradient Descent, Adam \cite{kingma2014adam}, and RMSprop \cite{mukkamala2017variants}, each of which can significantly impact the performance of a deep learning model. Thus, the selection of the appropriate optimizer is a critical consideration.

The MadGrad \cite{defazio2022adaptivity} optimizer, a variation of the commonly used Adam \cite{kingma2014adam}, aims to improve the generalization performance of deep learning models. This is achieved by incorporating multiplicative noise into the gradients during the training process. Specifically, the gradients are multiplied by a random noise tensor at each iteration, derived from a Gaussian distribution with a mean of 1 and a slight variance, and scaled by a factor that decreases over time. Results have demonstrated that MadGrad outperforms Adam on various tasks, including image classification and natural language processing. This study validated its effectiveness, resulting in a 1.1\% improvement in accuracy.

\subsection{Learning rate schedulers}
The objective of a learning rate scheduler in deep learning is to modulate the learning rate. This crucial hyperparameter determines the magnitude of updates made to the model's parameters by the optimizer. Ineffective regulation of the learning rate can result in either significant, erratic updates that lead to suboptimal convergence or slow, incremental updates that impede the pace of the training process. A learning rate scheduler resolves these issues by dynamically adjusting the learning rate during training, using various techniques such as fixed scheduling, scheduling based on the training progress, or scheduling based on model performance. This contribution demonstrates the efficacy of the combination of the Cosine Annealing scheduler \cite{cazenave2022cosine} and the ReducedLR scheduler \cite{al2022scheduling} in stabilizing the learning process.

The ReducedLR scheduler \cite{al2022scheduling} is a commonly utilized approach in deep learning for regulating the learning rate during the training process. This scheduler operates by reducing the learning rate by a predetermined factor when the training loss fails to improve over a specified number of epochs, thereby mitigating the risk of getting stuck in a suboptimal local minimum and enhancing the generalization performance of the model. Implementable as a callback function within deep learning frameworks, the ReducedLR scheduler offers the flexibility to specify the reduction factor and the number of epochs for which the learning rate reduction should be triggered.

The CosineAnnealing scheduler \cite{cazenave2022cosine} is a learning rate scheduler that adjusts the learning rate following a cosine curve, which commences at a high value and gradually decreases to a lower value as training progresses. This technique helps to improve the convergence of the training process and address the challenge of premature saturation, where the learning rate becomes excessively low and the training slows down.

\begin{table*}[]
\centering
\begin{tabular}{|c|c|c|c|} 
\hline
\multirow{2}{*}{Architecture} & \multirow{2}{*}{Technique} & \multicolumn{2}{l|}{NTU}  \\ 
\cline{3-4}
                              &                            & CS    & CV                \\ 
\hline
\multirow{4}{*}{RNN-based}    & S-trans+RNN                & 76.0  & 82.3              \\ 
\cline{2-4}
                              & S-trans+RNN (aug.)         & 77.0  & 85.0              \\ 
\cline{2-4}
                              & VA-RNN~                    & 79.4~ & 87.6              \\ 
\cline{2-4}
                              & VA-RNN (aug.) ~            & 79.8  & 88.9              \\ 
\hline
\multirow{5}{*}{CNN-Based}             & S-trans+CNN ~              & 87.5  & 92.2              \\ 
\cline{2-4}
                              & S-trans+CNN (aug.)~        & 87.9  & ~93.5             \\ 
\cline{2-4}
                              & VA-CNN ~                   & 88.2  & 93.8              \\ 
\cline{2-4}
                              & VA-CNN (aug.) ~            & 88.7  & 94.3              \\ 
\cline{2-4}
                              & Our + ArcFace              & -     & \textbf{95.0}              \\
\hline
\end{tabular}
	\caption{Effectiveness (in accuracy ($\%$)) of applying (regularlization, Madgrad Optimizer, multiple learning schedulers) }
	
	\label{fig:local_levelVSserver_level}
\end{table*}

\begin{table*}[]
\centering
\begin{tabular}{|c|c|c|c|}
\hline
Augmentation~         & Image-based & Weak  & Strong  \\ \cline{1-4}
Flipping              & Both        & 93.70 & 93.62   \\ \cline{1-4}
Rotation              & Both        & 94.21 & 94.1    \\ \cline{1-4}
Zoom                  & Image       & 93.60 & 93.40   \\ \cline{1-4}
Shear                 & Image       & 94.03 & 94.20   \\ \cline{1-4}
Translate-x           & Image       & 93.71 & 94.03   \\ \cline{1-4}
Translate-y           & Image       & 94.24 & 94.1    \\ \cline{1-4}
Cutout                & Image       & 94.03 & 94.1    \\ \cline{1-4}
Salt and pepper noise & Image       & 93.52 & -       \\ \cline{1-4}
Bone Shuffling        & Pose        & 94.13 & 93.96   \\ \cline{1-4}
Bone Masking          & Pose        & 93.80 & 93.8    \\ \cline{1-4}
Frame Masking         & Both        & 93.8  & 93.75   \\ \cline{1-4}
Gaussian Noise        & Image       & 93.5  & -       \\ \cline{1-4}
Speckle Noise         & Image       & 93.45 & -       \\ \cline{1-4}
Localvars Noise       & Image       & 93.5  & -       \\ \cline{1-4}
Salt noise            & Image       & 93.5  & -       \\ \cline{1-4}
Pepper Noise          & Image       & 93.5  & -     \\ \cline{1-4}
\end{tabular}
	\caption{Various augmentations applied on the encoded skeleton image map}
	
	\label{fig:local_levelVSserver_level}
\end{table*}

\section{Regularization}
Regularization is a widely employed technique in machine learning for mitigating overfitting, which refers to the scenario in which a model demonstrates superior performance on the training data yet subpar performance on unseen data. Overfitting transpires when the model has learned specific patterns in the training data that do not generalize to other data, resulting in poor performance on previously unseen data. Regularization helps to address overfitting by incorporating a penalty term in the loss function during the training process. This promotes the model to learn more generalizable parameters, thereby yielding enhanced performance on new data. There exist several types of regularization methods, such as label smoothing regularization, dropout, batch normalization, and early stopping, which can be utilized in a complementary manner to improve the generalizability of machine learning models.

Label Smoothing Regularization \cite{muller2019does} is a popular regularization technique utilized in deep learning to enhance the generalization performance of classifiers. The objective of label smoothing is to reduce the confidence of the classifier's predictions, thereby improving its generalization capabilities on new data. This technique is commonly utilized in image classification tasks, where instead of using one-hot encoded training data labels with a label vector of [0, 0, 1, 0] to indicate that an image belongs to class 3, label smoothing would modify the label vector to [0.1, 0.1, 0.8, 0.1], suggesting that the image has a high probability of belonging to class 3 but also a tiny likelihood of belonging to the other classes. This regularizes the classifier and reduces the risk of overfitting the training data. Essentially, one-hot encoded training data labels should not contain zero values for the non-class index.

Early stopping, as described in \cite{prechelt1998early}, is a prevalent regularization strategy utilized in deep learning to mitigate overfitting. The technique entails prematurely halting the training process prior to the model's convergence to its optimal performance. This is achieved by continuously monitoring the model's performance on a validation set and interrupting the training when the performance either ceases to improve or begins to decline. The objective of early stopping is to restrict overfitting, which is a situation where a model performs well on the training data but poorly on unseen data. By curtailing the training before the model attains full convergence, early stopping can enhance its ability to generalize to novel data and improve its overall performance. Early stopping is a straightforward and practical approach that can be effortlessly integrated into most deep-learning frameworks.

Dropout, as a regularization technique in deep learning, aims to enhance the generalization performance of neural networks \cite{srivastava2014dropout}. The approach involves randomly setting a specified fraction of the activations of the neurons in the network to zero during training. This reduction in the co-adaptation of neurons leads to a network that is less sensitive to the specific weights of individual neurons. As a result, training a network with dropouts can lead to learning more robust and diverse feature representations, which are less prone to overfitting the training data. Dropout is commonly applied to fully-connected layers in neural networks but can also be implemented in other layer types, such as convolutional and recurrent layers. It is usually deployed with a high dropout rate (e.g., 0.5) during training and a low dropout rate (e.g., 0.1) or no dropout during inference.

Batch normalization, as discussed in \cite{ioffe2015batch}, is a technique employed in deep learning to improve the performance and stability of neural networks. The strategy entails normalizing the inputs to each layer in the network, which can expedite the training process and enhance the generalization performance of the network. A batch normalization layer is inserted between the input and output of a neural network. It is designed to normalize the activations of the preceding layer using the mean and variance of the current mini-batch of data. This reduction of internal covariate shift, which refers to the phenomenon of the distribution of inputs to each layer evolving over time during training, can enhance the network's convergence speed and stability and make it more resilient to initialization parameter choices.

\section{Dataset and Performance Measures}
\subsection{Dataset}
The design of a data collection environment for capturing actions is dictated by the target task. For instance, the setup required for a gesture recognition task will differ from that required for a human interaction task. In most cases, the captured action occurs at the beginning of the sequence, obviating the need for action recognition without action detection. Additionally, many datasets use a global capturing length, which simplifies the data preprocessing by eliminating the requirement for consistent length. The presence of non-action-related subjects in the scene is another challenge commonly addressed by capturing data in a controlled environment to ensure that only sequences relevant to the task are captured.

The NTU RGB+D dataset, proposed by Shahroudy \textit{et al.} \cite{shahroudy2016ntu}, is a substantial multi-view action dataset with 56,000 samples from 40 individuals and 60 action classes, which are classified into three categories: 40 daily activity classes, nine health-related classes, and 11 interaction classes. The Microsoft Kinect device was used for data collection, providing RGB, depth, skeleton, and infrared modalities. Three fixed camera setups were utilized, with capturing angles ranging from -45 to 45 degrees, and the cameras' distance and height were also varied to increase view variations. Liu \textit{et al.} \cite{liu2019ntu} further extended the NTU-60 dataset using the same capturing system and modalities, with 106 subjects performing 120 action classes and 114,500 video samples.

\subsection{\textbf{Performance Measures}}
Standardizing evaluation metrics is crucial for fairly comparing various approaches. The choice of metric depends on the task, and some metrics may be more intuitive than others. The choice of evaluation protocol may also play a role in demonstrating the difficulty and complexity of the scenario. The \textbf{Cross-views} evaluation protocol, proposed by Shahroudy \textit{et al.} \cite{shahroudy2016ntu} and Liu \textit{et al.} \cite{liu2017pku}, assumes that samples from the same view cannot be used for both training and testing, for instance, camera 1 and 3 for training and camera 2 for testing only. 

\textbf{Cross-subject} evaluation protocol, also proposed by the creators of the NTU-D 60 dataset \cite{shahroudy2016ntu, liu2017pku}, dictates that the subjects selected for training cannot be used for testing, with 20 of the 40 subjects selected for training and the remaining 20 for testing.
Lastly, Liu \textit{et al.} \cite{liu2017pku} also proposed the \textbf{cross-setup} evaluation protocol, which uses different vertical heights, distances, and backgrounds during the capturing to include natural variations, while keeping the horizontal three-camera fixed in terms of capturing angle.

\section{Results}
We replicated the results of the End-to-End Two Streams Network for Action Recognition as reported by \textit{et al.} \cite{zhang2019view}. We established it as a baseline for our experiments. In their work, Zhang et al. \textit{et al.} \cite{zhang2019view} built upon their previous contribution in \cite{zhang2017view} by proposing an End-to-End Two Streams Network for Action Recognition. The network comprises two streams: one is similar to the one presented in \cite{zhang2017view}, while the other is a CNN-based stream that includes a view adaptation subnetwork with a similar architecture. This CNN stream leverages the skeleton representation proposed by Du \textit{et al.} \cite{du2015skeleton}. To enhance the robustness of the network, random rotation augmentation was incorporated on-the-fly. Finally, late score fusion was applied with different stream weights, favoring the CNN stream. Our results outperform those of the Two-Stream View Adaptive Module, demonstrating the potential of our approach with simple training techniques. We incorporated all mentioned strategies and techniques to achieve the results shown in the tables.

\section{Conclusion}
The proliferation of sensing devices has increased the availability of diverse datasets in multiple modalities. Among these, skeleton-based modality holds promise due to its computation efficiency and the wealth of information the human skeleton provides. Accurately estimating human pose from the video is a prerequisite for extracting 3D skeleton data from various modalities.
In this work, we have shown that Convolutional Neural Networks (CNNs) utilizing diverse training techniques can achieve state-of-the-art (SOTA) results comparable to those obtained using Graph Neural Networks (GNNs) for action recognition tasks. Furthermore, our findings indicate that using various data augmentation techniques can improve the generalization and robustness of the model. This results in improved performance on unseen data and reduced sensitivity to variations or distortions in the input.
Additionally, we have demonstrated that using MadGrad as the optimizer and implementing a learning rate scheduler can improve the model's accuracy. Furthermore, we have shown that using a margin-based cosine loss function instead of the traditional cross-entropy loss can enhance the model's performance, leading to more accurate predictions and improved overall results. Regularization techniques can further prevent overfitting, improving the model's performance on unseen data.

\bibliography{main}

\begin{thebibliography}{10}

\bibitem{krizhevsky2012imagenet}
Alex Krizhevsky, Ilya Sutskever, and Geoffrey~E Hinton.
\newblock Imagenet classification with deep convolutional neural networks.
\newblock {\em Advances in neural information processing systems},
  25:1097--1105, 2012.

\bibitem{ren2015faster}
Shaoqing Ren, Kaiming He, Ross Girshick, and Jian Sun.
\newblock Faster r-cnn: Towards real-time object detection with region proposal
  networks.
\newblock {\em Advances in neural information processing systems}, 28:91--99,
  2015.

\bibitem{barker2006accuracy}
Susan Barker, Rebecca Craik, William Freedman, Nira Herrmann, and Howard
  Hillstrom.
\newblock Accuracy, reliability, and validity of a spatiotemporal gait analysis
  system.
\newblock {\em Medical engineering \& physics}, 28(5):460--467, 2006.

\bibitem{huang2020long}
Junqin Huang, Xiang Xiang, Xuan Gong, Baochang Zhang, et~al.
\newblock Long-short graph memory network for skeleton-based action
  recognition.
\newblock In {\em Proceedings of the IEEE/CVF Winter Conference on Applications
  of Computer Vision}, pages 645--652, 2020.

\bibitem{raj2020exploring}
Bharath~N Raj, Anand Subramanian, Kashyap Ravichandran, and Dr~N Venkateswaran.
\newblock Exploring techniques to improve activity recognition using human pose
  skeletons.
\newblock In {\em Proceedings of the IEEE/CVF Winter Conference on Applications
  of Computer Vision Workshops}, pages 165--172, 2020.

\bibitem{huynh2009metrics}
Du~Q Huynh.
\newblock Metrics for 3d rotations: Comparison and analysis.
\newblock {\em Journal of Mathematical Imaging and Vision}, 35(2):155--164,
  2009.

\bibitem{yan2019convolutional}
Sijie Yan, Zhizhong Li, Yuanjun Xiong, Huahan Yan, and Dahua Lin.
\newblock Convolutional sequence generation for skeleton-based action
  synthesis.
\newblock In {\em Proceedings of the IEEE/CVF International Conference on
  Computer Vision}, pages 4394--4402, 2019.

\bibitem{li2018co}
Chao Li, Qiaoyong Zhong, Di~Xie, and Shiliang Pu.
\newblock Co-occurrence feature learning from skeleton data for action
  recognition and detection with hierarchical aggregation.
\newblock {\em arXiv preprint arXiv:1804.06055}, 2018.

\bibitem{boser1992training}
Bernhard~E Boser, Isabelle~M Guyon, and Vladimir~N Vapnik.
\newblock A training algorithm for optimal margin classifiers.
\newblock In {\em Proceedings of the fifth annual workshop on Computational
  learning theory}, pages 144--152, 1992.

\bibitem{liu2017fusing}
Mengyuan Liu, Qinqin He, and Hong Liu.
\newblock Fusing shape and motion matrices for view invariant action
  recognition using 3d skeletons.
\newblock In {\em 2017 IEEE International Conference on Image Processing
  (ICIP)}, pages 3670--3674. IEEE, 2017.

\bibitem{weng2017spatio}
Junwu Weng, Chaoqun Weng, and Junsong Yuan.
\newblock Spatio-temporal naive-bayes nearest-neighbor (st-nbnn) for
  skeleton-based action recognition.
\newblock In {\em Proceedings of the IEEE Conference on computer vision and
  pattern recognition}, pages 4171--4180, 2017.

\bibitem{tang2014human}
Nick~C Tang, Yen-Yu Lin, Ju-Hsuan Hua, Ming-Fang Weng, and Hong-Yuan~Mark Liao.
\newblock Human action recognition using associated depth and skeleton
  information.
\newblock In {\em 2014 IEEE International Conference on Acoustics, Speech and
  Signal Processing (ICASSP)}, pages 4608--4612. IEEE, 2014.

\bibitem{ubalde2016skeleton}
Sebasti{\'a}n Ubalde, Francisco G{\'o}mez-Fern{\'a}ndez, Norberto~A Goussies,
  and Marta Mejail.
\newblock Skeleton-based action recognition using citation-knn on bags of
  time-stamped pose descriptors.
\newblock In {\em 2016 IEEE International Conference on Image Processing
  (ICIP)}, pages 3051--3055. IEEE, 2016.

\bibitem{duan2021revisiting}
Haodong Duan, Yue Zhao, Kai Chen, Dian Shao, Dahua Lin, and Bo~Dai.
\newblock Revisiting skeleton-based action recognition.
\newblock {\em arXiv preprint arXiv:2104.13586}, 2021.

\bibitem{wang2021proposal}
Xiang Wang, Zhiwu Qing, Ziyuan Huang, Yutong Feng, Shiwei Zhang, Jianwen Jiang,
  Mingqian Tang, Changxin Gao, and Nong Sang.
\newblock Proposal relation network for temporal action detection.
\newblock {\em arXiv preprint arXiv:2106.11812}, 2021.

\bibitem{du2015skeleton}
Yong Du, Yun Fu, and Liang Wang.
\newblock Skeleton based action recognition with convolutional neural network.
\newblock In {\em 2015 3rd IAPR Asian conference on pattern recognition
  (ACPR)}, pages 579--583. IEEE, 2015.

\bibitem{wang2016action}
Pichao Wang, Zhaoyang Li, Yonghong Hou, and Wanqing Li.
\newblock Action recognition based on joint trajectory maps using convolutional
  neural networks.
\newblock In {\em Proceedings of the 24th ACM international conference on
  Multimedia}, pages 102--106, 2016.

\bibitem{li2017joint}
Chuankun Li, Yonghong Hou, Pichao Wang, and Wanqing Li.
\newblock Joint distance maps based action recognition with convolutional
  neural networks.
\newblock {\em IEEE Signal Processing Letters}, 24(5):624--628, 2017.

\bibitem{li2017skeleton}
Chao Li, Qiaoyong Zhong, Di~Xie, and Shiliang Pu.
\newblock Skeleton-based action recognition with convolutional neural networks.
\newblock In {\em 2017 IEEE International Conference on Multimedia \& Expo
  Workshops (ICMEW)}, pages 597--600. IEEE, 2017.

\bibitem{ke2017skeletonnet}
Qiuhong Ke, Senjian An, Mohammed Bennamoun, Ferdous Sohel, and Farid Boussaid.
\newblock Skeletonnet: Mining deep part features for 3-d action recognition.
\newblock {\em IEEE signal processing letters}, 24(6):731--735, 2017.

\bibitem{du2015hierarchical}
Yong Du, Wei Wang, and Liang Wang.
\newblock Hierarchical recurrent neural network for skeleton based action
  recognition.
\newblock In {\em Proceedings of the IEEE conference on computer vision and
  pattern recognition}, pages 1110--1118, 2015.

\bibitem{du2016representation}
Yong Du, Yun Fu, and Liang Wang.
\newblock Representation learning of temporal dynamics for skeleton-based
  action recognition.
\newblock {\em IEEE Transactions on Image Processing}, 25(7):3010--3022, 2016.

\bibitem{veeriah2015differential}
Vivek Veeriah, Naifan Zhuang, and Guo-Jun Qi.
\newblock Differential recurrent neural networks for action recognition.
\newblock In {\em Proceedings of the IEEE international conference on computer
  vision}, pages 4041--4049, 2015.

\bibitem{song2017end}
Sijie Song, Cuiling Lan, Junliang Xing, Wenjun Zeng, and Jiaying Liu.
\newblock An end-to-end spatio-temporal attention model for human action
  recognition from skeleton data.
\newblock In {\em Proceedings of the AAAI conference on artificial
  intelligence}, 2017.

\bibitem{zhang2017geometric}
Songyang Zhang, Xiaoming Liu, and Jun Xiao.
\newblock On geometric features for skeleton-based action recognition using
  multilayer lstm networks.
\newblock In {\em 2017 IEEE Winter Conference on Applications of Computer
  Vision (WACV)}, pages 148--157. IEEE, 2017.

\bibitem{yan2018spatial}
Sijie Yan, Yuanjun Xiong, and Dahua Lin.
\newblock Spatial temporal graph convolutional networks for skeleton-based
  action recognition.
\newblock In {\em Thirty-second AAAI conference on artificial intelligence},
  2018.

\bibitem{shi2020skeleton}
Lei Shi, Yifan Zhang, Jian Cheng, and Hanqing Lu.
\newblock Skeleton-based action recognition with multi-stream adaptive graph
  convolutional networks.
\newblock {\em IEEE Transactions on Image Processing}, 29:9532--9545, 2020.

\bibitem{song2019richly}
Yi-Fan Song, Zhang Zhang, and Liang Wang.
\newblock Richly activated graph convolutional network for action recognition
  with incomplete skeletons.
\newblock In {\em 2019 IEEE International Conference on Image Processing
  (ICIP)}, pages 1--5. IEEE, 2019.

\bibitem{yoon2021predictively}
Yongsang Yoon, Jongmin Yu, and Moongu Jeon.
\newblock Predictively encoded graph convolutional network for noise-robust
  skeleton-based action recognition.
\newblock {\em Applied Intelligence}, pages 1--15, 2021.

\bibitem{de2020infrared}
Alban~Main De~Boissiere and Rita Noumeir.
\newblock Infrared and 3d skeleton feature fusion for rgb-d action recognition.
\newblock {\em IEEE Access}, 8:168297--168308, 2020.

\bibitem{lee2017ensemble}
Inwoong Lee, Doyoung Kim, Seoungyoon Kang, and Sanghoon Lee.
\newblock Ensemble deep learning for skeleton-based action recognition using
  temporal sliding lstm networks.
\newblock In {\em Proceedings of the IEEE international conference on computer
  vision}, pages 1012--1020, 2017.

\bibitem{rahmani2017learning}
Hossein Rahmani and Mohammed Bennamoun.
\newblock Learning action recognition model from depth and skeleton videos.
\newblock In {\em Proceedings of the IEEE International Conference on Computer
  Vision}, pages 5832--5841, 2017.

\bibitem{zhang2019view}
Pengfei Zhang, Cuiling Lan, Junliang Xing, Wenjun Zeng, Jianru Xue, and Nanning
  Zheng.
\newblock View adaptive neural networks for high performance skeleton-based
  human action recognition.
\newblock {\em IEEE transactions on pattern analysis and machine intelligence},
  41(8):1963--1978, 2019.

\bibitem{cubuk2020randaugment}
Ekin~D Cubuk, Barret Zoph, Jonathon Shlens, and Quoc~V Le.
\newblock Randaugment: Practical automated data augmentation with a reduced
  search space.
\newblock In {\em Proceedings of the IEEE/CVF conference on computer vision and
  pattern recognition workshops}, pages 702--703, 2020.

\bibitem{rao2021augmented}
Haocong Rao, Shihao Xu, Xiping Hu, Jun Cheng, and Bin Hu.
\newblock Augmented skeleton based contrastive action learning with momentum
  lstm for unsupervised action recognition.
\newblock {\em Information Sciences}, 569:90--109, 2021.

\bibitem{devries2017improved}
Terrance DeVries and Graham~W Taylor.
\newblock Improved regularization of convolutional neural networks with cutout.
\newblock {\em arXiv preprint arXiv:1708.04552}, 2017.

\bibitem{shorten2019survey}
Connor Shorten and Taghi~M Khoshgoftaar.
\newblock A survey on image data augmentation for deep learning.
\newblock {\em Journal of big data}, 6(1):1--48, 2019.

\bibitem{liu2022towards}
Xiaogeng Liu, Haoyu Wang, Yechao Zhang, Fangzhou Wu, and Shengshan Hu.
\newblock Towards efficient data-centric robust machine learning with
  noise-based augmentation.
\newblock {\em arXiv preprint arXiv:2203.03810}, 2022.

\bibitem{suarez2021tutorial}
Juan~Luis Su{\'a}rez, Salvador Garc{\'\i}a, and Francisco Herrera.
\newblock A tutorial on distance metric learning: Mathematical foundations,
  algorithms, experimental analysis, prospects and challenges.
\newblock {\em Neurocomputing}, 425:300--322, 2021.

\bibitem{deng2019arcface}
Jiankang Deng, Jia Guo, Niannan Xue, and Stefanos Zafeiriou.
\newblock Arcface: Additive angular margin loss for deep face recognition.
\newblock In {\em Proceedings of the IEEE/CVF conference on computer vision and
  pattern recognition}, pages 4690--4699, 2019.

\bibitem{kingma2014adam}
Diederik~P Kingma and Jimmy Ba.
\newblock Adam: A method for stochastic optimization.
\newblock {\em arXiv preprint arXiv:1412.6980}, 2014.

\bibitem{mukkamala2017variants}
Mahesh~Chandra Mukkamala and Matthias Hein.
\newblock Variants of rmsprop and adagrad with logarithmic regret bounds.
\newblock In {\em International conference on machine learning}, pages
  2545--2553. PMLR, 2017.

\bibitem{defazio2022adaptivity}
Aaron Defazio and Samy Jelassi.
\newblock Adaptivity without compromise: a momentumized, adaptive, dual
  averaged gradient method for stochastic optimization.
\newblock {\em Journal of Machine Learning Research}, 23:1--34, 2022.

\bibitem{cazenave2022cosine}
Tristan Cazenave, Julien Sentuc, and Mathurin Videau.
\newblock Cosine annealing, mixnet and swish activation for computer go.
\newblock In {\em Advances in Computer Games}, pages 53--60. Springer, 2022.

\bibitem{al2022scheduling}
Ayman Al-Kababji, Faycal Bensaali, and Sarada~Prasad Dakua.
\newblock Scheduling techniques for liver segmentation: Reducelronplateau vs
  onecyclelr.
\newblock {\em arXiv preprint arXiv:2202.06373}, 2022.

\bibitem{muller2019does}
Rafael M{\"u}ller, Simon Kornblith, and Geoffrey~E Hinton.
\newblock When does label smoothing help?
\newblock {\em Advances in neural information processing systems}, 32, 2019.

\bibitem{prechelt1998early}
Lutz Prechelt.
\newblock Early stopping-but when?
\newblock In {\em Neural Networks: Tricks of the trade}, pages 55--69.
  Springer, 1998.

\bibitem{ioffe2015batch}
Sergey Ioffe and Christian Szegedy.
\newblock Batch normalization: Accelerating deep network training by reducing
  internal covariate shift.
\newblock In {\em International conference on machine learning}, pages
  448--456. PMLR, 2015.

\bibitem{shahroudy2016ntu}
Amir Shahroudy, Jun Liu, Tian-Tsong Ng, and Gang Wang.
\newblock Ntu rgb+ d: A large scale dataset for 3d human activity analysis.
\newblock In {\em Proceedings of the IEEE conference on computer vision and
  pattern recognition}, pages 1010--1019, 2016.

\bibitem{liu2019ntu}
Jun Liu, Amir Shahroudy, Mauricio Perez, Gang Wang, Ling-Yu Duan, and Alex~C
  Kot.
\newblock Ntu rgb+ d 120: A large-scale benchmark for 3d human activity
  understanding.
\newblock {\em IEEE transactions on pattern analysis and machine intelligence},
  42(10):2684--2701, 2019.

\bibitem{liu2017pku}
Chunhui Liu, Yueyu Hu, Yanghao Li, Sijie Song, and Jiaying Liu.
\newblock Pku-mmd: A large scale benchmark for continuous multi-modal human
  action understanding.
\newblock {\em arXiv preprint arXiv:1703.07475}, 2017.

\bibitem{zhang2017view}
Pengfei Zhang, Cuiling Lan, Junliang Xing, Wenjun Zeng, Jianru Xue, and Nanning
  Zheng.
\newblock View adaptive recurrent neural networks for high performance human
  action recognition from skeleton data.
\newblock In {\em Proceedings of the IEEE International Conference on Computer
  Vision}, pages 2117--2126, 2017.

\end{thebibliography}

\end{document}